\documentclass[11pt,letterpaper]{article}
\usepackage{emnlp2016}
\usepackage{times}
\usepackage{latexsym}
\usepackage{times}
\usepackage{helvet}
\usepackage{courier}
\usepackage{url}
\usepackage{soul}
\usepackage{graphicx,float,wrapfig}
\usepackage{multirow}
\usepackage{verbatim}

\emnlpfinalcopy

\def\emnlppaperid{863}

\title{Predicting the Relative Difficulty of Single Sentences \\ With and Without Surrounding Context}

\author{Elliot Schumacher \\ Carnegie Mellon University \\ eschumac@cs.cmu.edu 
\And
Maxine Eskenazi \\ Carnegie Mellon  University \\ max@cs.cmu.edu
\AND
Gwen Frishkoff \\ University of Oregon \\ gfrishkoff@gmail.com 
\And
Kevyn Collins-Thompson \\ University of Michigan \\ kevynct@umich.edu}

\date{}

\begin{document}

\maketitle
  
\begin{abstract}
  The problem of accurately predicting relative reading difficulty across a set of sentences arises in a number of important natural language applications, such as finding and curating effective usage examples for intelligent language tutoring systems. Yet while significant research has explored document- and passage-level reading difficulty, the special challenges involved in assessing aspects of readability for single sentences have received much less attention, particularly when considering the role of surrounding passages. We introduce and evaluate a novel approach for estimating the relative reading difficulty of a set of sentences, with and without surrounding context. Using different sets of lexical and grammatical features, we explore models for predicting pairwise relative difficulty using logistic regression, and examine rankings generated by aggregating pairwise difficulty labels using a Bayesian rating system to form a final ranking. We also compare rankings derived for sentences assessed with and without context, and find that contextual features can help predict differences in relative difficulty judgments across these two conditions.
\end{abstract}

\section{Introduction}
The reading difficulty, or \emph{readability}, of a text is an estimate of linguistic complexity and is typically based on lexical and syntactic features, such as text length, word frequency, and grammatical complexity \cite{collins2004language,schwarm2005reading,kidwell2011statistical,kanungo2009predicting}. Such estimates are often expressed as age- or grade-level measures and are useful for a range of educational and research applications. For example, instructors often wish to select stories or books that are appropriately matched to student grade level. 

Many measures have been designed to calculate readability at the document level (e.g., for web pages, articles, or books)   \cite{collins2004language,schwarm2005reading}, as well as the paragraph or passage level \cite{kidwell2011statistical,kanungo2009predicting}. However, much less work has attempted to characterize the readability of single sentences~\cite{pilan2014rule}. This problem is challenging because single sentences provide less data than is typically required for reliable estimates, particularly for measures that rely on aggregate statistics.

The absence of reliable single-sentence estimates points to a gap in natural language processing (NLP) research. Single sentences are used in a variety of experimental and NLP applications: for example, in studies of reading comprehension. Because readability estimates have been shown to predict a substantial portion of variance in comprehension of different texts, it would be  useful to have measures of single-sentence readability. Thus, one aim of the current study was to estimate the \emph{relative readability} of single sentences with a high degree of accuracy. To our knowledge, general-purpose methods for computing such estimates for native language (L1) readers have not been developed, and thus our goal was to develop a method that would characterize sentence-level difficulty for that group.

The second aim is to compare the readability of single sentences in isolation with the readability of these same sentences embedded in a larger context (e.g., paragraph, passage, or document). When a single sentence is extracted from a text, it is likely to contain linguistic elements, such as anaphora (e.g., ``he" or ``the man"), that are semantically or syntactically dependent on surrounding context. Not surprisingly, sentences that contain these contextual dependencies take more effort to comprehend: an anaphoric noun phrase, or NP (e.g., ``he"), automatically triggers the need to resolve reference, typically by understanding the link between the anaphor and a full NP from a previous sentence (e.g., ``John" or ``The man that I introduced you to at the party last night"~\cite{frishkoffperfetti2008}. In general, studies have shown a link between reading comprehension and the presence of such cross-sentence relationships in the text~\cite{mcnamara2001reading,liederholm2000effects,vosscomprehension1996}.  This implies that the very notion of readability at the sentence level may depend on context as well as word- and sentence-level features. Therefore, it is important to compare readability estimates for single sentences that occur in isolation with those that occur within a larger passage, particularly if the target sentence contains coreferences, implied meanings, or other dependencies with its context.

To address these aims, the present study first conducted two crowdsourcing experiments. In the first, `sentence-only' experiment, workers were asked to judge which of two ``target" sentences they thought was more difficult. In the second, `sentence-in-passage' experiment, another group of workers was presented with the same target sentences that were used in the first experiment. However, in the second experiment, target sentences were embedded in their original contexts. 

Next, we analyzed these judgments of relative readability for each condition (sentence-only versus sentence-in-passage) by developing models for predicting pairwise relative difficulty of sentences. These models used a rich representation of target sentences based on a combination of lexical, syntactic, and discourse features.  Significant differences were found in readability judgments for sentences with and without their surrounding context. This demonstrates that discourse-level features (i.e., features related to coherence and cohesion) can affect the readability of single sentences.



\section{Related Work}

Recent approaches to estimating readability have used a variety of linguistic features and prediction models~\cite{collins2014survey}.  The Lexile Framework \cite{stenner1996measuring} uses word frequency estimates in a large corpus as a proxy for lexical difficulty, and sentence length as a grammatical feature. Methods based on statistical machine learning, such as the reading difficulty measures developed by Collins-Thompson and Callan~\cite{collins2004language} 
and \cite{schwarm2005reading} used a feature set based on language models. Later work \cite{Heilman:2008:ASM:1631836.1631845} incorporated grammatical features by parsing the sentences in a text, and creating subtrees of one- to three-level depth as separate features. Such features allow more detailed, direct analysis of the sentence structure itself instead of  traditional proxies for syntactic complexity likes sentence length. 
The linguistic features proposed in these works capture specific aspects of language difficulty applied at the document level, whereas our work investigates the effectiveness of these feature types for characterizing aspects of difficulty at the sentence level.

Methods have been proposed to measure the readability of shorter portions of text (e.g. typically less than 100 words), including sentences.  The approach most similar to ours is the prediction of relative sentence difficulty (with associated readability ranking) for the deaf introduced by Inui et al. (2001)\nocite{InuiY01}.  That work focused on effective morphosyntactic features for that target population with an SVM binary classifier, whereas our approach (1) is intended for a broader population of L1 learners and thus explores the effectiveness of adding a rich, lexically-derived feature set, (2) uses a logistic regression model to estimate class probabilities and interprets the results of that model, compared to applying an SVM without interpretation to obtain a binary label, (3) examines differences in predicting sentence difficulty both in and out of passage context, and (4) creates and uses a new dataset based on a crowdsourced approach, using hundreds of non-experts to gather thousands of pairwise preferences, compared to a questionnaire deployed to a small number of experts. In other domains, a model was proposed to predict the readability of short web summaries in Kanungo and Orr 2009\nocite{kanungo2009predicting}.  In Kidwell et al. (2011), \nocite{kidwell2011statistical} a set of Age of Acquisition estimates for individual words, representing the lexical component of difficulty, was used to predict the grade levels of passages. Some approaches have explored the classification of specific aspects of sentences, as opposed to reading difficulty classification.  For example, \cite{pilan2014rule} classified individual sentences that would be understood by second-language learners. Another work \cite{kilgarriff2008gdex} identified sentences that would be good dictionary examples by looking for specific desirable features. Davenport et al. 2014 \nocite{DBLP:journals/corr/DavenportD14} used a traditional method of readability (Flesch-Kincaid), within the larger context of exploring relationships between the difficulty of tweets in a geographic area and demographics. Research in text simplification has applied sentence-level models of difficulty as part of simplification-based optimization objectives. For example, Woodsend and Lapata (2011)\nocite{Woodsend:2011} use word and syllable count as proxy features for sentence difficulty when implicitly comparing different simplified variants of a sentence.


Other approaches have considered the relationship of reading difficulty to structures within in the whole text.  These relationships can include the number of coreferences present in a text. Coh-Metrix \cite{graesser2011coh} measures text cohesiveness, accounting for both the reading difficulty of the text and other lexical and syntactic measures as well as a measure of prior knowledge needed for comprehension, and the genre of the text. Coh-Metrix uses co-reference detection as a factor in the cohesiveness of a text, typically at the document or passage level. Such cohesiveness factors account for the difficulty of constructing the mental representation of texts with more complex internal structure.  TextEvaluator \cite{sheehan2013two,sheehan2014textevaluator} is designed to help educators select materials for instruction.  The tool includes several components in its evaluation of text, including narrativity, style, and cohesion, beyond traditional difficulty and is again at the whole document level.  This approach illustrates that the difficulty of a text relies on the relationships within it.  This motivates the need to consider context when measuring difficulty.

Generating reading difficulty rankings of longer texts from pairwise preferences has been performed in other contexts.  Tanaka-Ishii et al. (2010) \nocite{TanakaIshii:2010} explored an approach for sorting texts by readability based on pairwise preferences. Later, Chen et al. (2013) \nocite{chen2013pairwise} also proposed a model to obtain passage readability ranking by aggregating pairwise comparisons made by crowdworkers.  
In De Clercq et al.(2014)\nocite{de2014using}, pairwise judgments of whole passages were obtained from crowdworkers and were found to give comparable results in aggregate to those obtained from experts. A pairwise ranking of text readability was created in Pitler and Nenkova (2008) \nocite{Pitler:2008:RRU:1613715.1613742} in which readability was defined by subjective questions asked to the reader after finishing the article, such as ``How well-written is this article?". All of the above previous work was focused on ordering longer text passages, not single sentences as we do here.

Finally, research in the Machine Translation field has explored pairwise prediction of the best translation between two sentence options.  For example, in Song and Cohn (2011)\nocite{song2011regression}, a pairwise prediction model was built using n-gram precision and recall, as well as  function, content, and word counts.  However, unlike pairwise prediction of difficulty, the prediction is done with respect to a reference sentence, or set of reference sentences.

\section{Data Collection and Processing}

We now describe methods used to create our dataset of sentences, to collect pairwise assessments of difficulty, and to aggregate these pairwise preferences into a complete ranking.

\subsection{Data Set}

The study sentences were drawn from a corpus combining the American National Corpus \cite{reppen2005american}, the New York Times Corpus \cite{sandhaus2008new}, and the North American News Text Corpus \cite{mcclosky2008bllip}. The domain of these corpora is largely news text, but also includes other topics, such as travel guides and other non-fiction.  In total, this database contains 60,663,803 sentences that served as initial candidates.  
Sentences were filtered out that didn't include one of the 70 target words that the third author selected for a study on teaching vocabulary to 8-14 year-old students.  Other sentences were removed based on length, keeping only sentences of between 6 and 20 words. Some sentences were removed due to the presence of one or more rare words. Finally, sentences were annotated with the surrounding document reading level, using a lexical readability model \cite{collins2004language}.
The data set gathered by \cite{collins2004language} was used in order to add to the amount of lower-level reading material in the collected corpora.

With these sentences, two crowdsourced tasks were prepared to gather pairwise assessments of sentence reading difficulty. In one task, the sentences were presented alone, outside of their original passage context. In the other task, the same sentences were presented within their original passage context. The objective was to generate two sets of pairwise comparisons of the readability of a sentence. In total, 120 sentence pairs were used for the first task and 120 passage pairs were used for the second. Each sentence was compared to five others, which created 300 comparisons in each task. The five sentences matched to each sentence were selected to ensure that pairs with a range of document level differences would be created. Within each type of pair, a random pair was selected.

There were several constraints when generating pairs for comparison. To allow for sentences to be taken from documents with a range of reading levels, sentences were selected evenly from documents at each reading level. From the twelve standard U.S. grade levels used in readability, each document was considered to be part of a bin consisting of two adjacent grade levels, such as grades 1 and 2, for example. Sentences were selected evenly from those bins.

Each sentence needed sufficient context to ensure there would be equivalent context for each item that would be compared, so only passages of sufficient size were included. To ensure passages were of similar length, only passages that had between 136 and 160 words were included. 
Contexts having at least two sentences before and after the sentence in question were strongly preferred. Each selected sentence was paired with one sentence from each of the other grade level bins. For example, a sentence from grade 1 would be paired with one sentence each from grade 3-4, 5-6, 7-8, 9-10, and 11-12. Finally, each pair of sentences was presented in AB and BA order. For each pair, there were seven worker decisions. There were 296 unique workers for the sentence-only task, and 355 for the sentence-in-passage task. 

\subsection{Crowdsourcing} 

Both of these tasks were carried out on the Crowdflower platform. The workers were first given instructions for each task, which included a description of the general purpose of the task. In the sentence-only task, workers were asked to select which of the two sentences was more difficult. In the sentence-within-passage task, workers were similarly asked to decide which underlined sentence was more difficult. The instructions for the latter requested that the workers make their judgment only on the sentence, not on the whole context. In both tasks, there was an option for ``I don't know or can't decide". The workers were asked to make their decision based on the vocabulary and grammatical structure of the sentences. Finally, examples for each task were provided with explanations for each answer. 

For each task, at least 40 gold standard questions were created from pairs of sentences that were judged to be sufficiently distinct from one another so that they could easily be answered correctly. For the sentence-in-passage task, several gold standard questions were written to verify that the instructions were being followed, since it was possible that a worker might judge the sentences based on the quality of the passage alone. These gold examples consisted of an easier sentence in a difficult passage compared with a difficult sentence within an easy passage. For each task, the worker saw three questions, including one gold standard question. A worker needed to maintain an 85\% accuracy rating on gold standard questions to continue, and needed to spend at least 25 seconds per page, which contained 3 questions each.

A weighted disagreement rate was calculated for each worker. If a worker's response to a question differed from the most frequent answer to that question, the percentage of agreement was counted against the worker.  If a worker, for the sentence-only task, had a disagreement rate (the weighted disagreement penalty divided by the total questions they answered) of 15\% or higher, their contribution was removed from the data set (or 17\% or higher for the sentence in passage task).  The agreement for the sentence-in-passage task is lower than the sentence-only task (88.93\% and 90.33\% respectively), so the permitted disagreement level is higher for that task.  This resulted in the removal of 5.7\% and 4.5\% of pairwise judgments, respectively. For each question, there was an optional text form to allow workers to submit feedback. The sentence-only task paid 11 cents per page, and the sentence-in-passage task paid 22 cents per page. 

\subsection{Ranking Generation} \label{rankingcreation}

Each task resulted in 4,200 pairwise preference judgments, excluding gold-standard answers.  To aggregate these pairwise preferences into an overall ranking of sentences, we use a simple, publicly available approach evaluated by Chen et al.~\nocite{chen2013pairwise} as being competitive with their own Crowd-BT aggregation method: the Microsoft Trueskill algorithm \cite{trueskill2007}. Trueskill is a Bayesian skill rating system that
generalized the well-known Elo rating system, in that it generates a ranking from pairwise decisions.
As Trueskill's ranking algorithm depends on the order in which the samples are processed, we report the ranking as an average of 50 runs.

The judgments were not aggregated for each comparison. Instead, each of the judgments was treated individually. This allows Trueskill to consider the degree of agreement between workers, since a sentence judgment that has high agreement reflects a larger difference in ranking than one that has lower agreement. Each sentence was considered a player, and the winner between two, A or B, was the sentence considered most difficult. If a worker chose ``I don't know or can't tell", it was considered a draw. The prediction resulting in ``I don't know or can't tell" is rare; 2.2\% of decisions in the sentence only task resulted in a draw, and 2.0\% for sentences within passages. After processing each of the judgments, a rating can be built of sentences, ranked from least difficult to most difficult. We can compare the resulting rankings for the sentence-only task and the sentence-in-passage task to see the effect of context on relative sentence difficulty. 

\section{Modeling Pairwise Relative Difficulty} \label{modelingsection}

Our first step in exploring relative difficulty ordering for a set of sentences was to develop a model that could accurately predict relative difficulty for a single pair of sentences, corresponding to the pairwise judgements of relative difficulty we gathered from the crowd. We did this for both the sentence-only and the sentence-in-passage tasks. In predicting a pairwise judgment for the sentence-only task, the model uses only the sentence texts. In the model for the sentence-in-passage task, the Stanford Deterministic Coreference Resolution System  \cite{raghunathan2010multi} is used to find coreference chains within the passage. From these coreference chains, sentences with references to and from the target sentence can be identified. If any additional sentences are found, these are used in a separate feature set that is included in the model; for all possible features, they are calculated for the target sentence, and separately for the additional sentence set.

Prior to training the final model, feature selection was done on random splits of the training data. Training data was used to fit a Random Forest Classifier, and based on the resulting classifier, the most important variables were selected using sklearn's feature importance method.  The top 2\% of the features (or 1\% for the sentence-in-passage with coreference, since the feature set size is doubled) were selected automatically.  This resulted in a feature size of 40-50 features.
We implemented our models using scikit-learn~\cite{scikit-learn} in Python.  

The resulting features were used to train a Logistic Regression model.  While other prediction models such as Support Vector Machines have been applied to relative readability prediction~\cite{InuiY01}, we chose Logistic Regression due to its ability to provide estimates of class probabilities (which may be important for reliability when deploying a system that recommends high-quality items for learners), its connection to the Rasch psychometric model used with reading assessments~\cite{EharaSON12}, and the interpretable nature of the resulting parameter weights.  
Since a given feature has a value for sentence A and B, if a feature was selected for only Sentence A or B, the feature for the other sentence was also added. We used the NLTK library \cite{bird2009natural} to tokenize the sentence for feature processing.

At the sentence level, the familiarity of the words is a significant factor to consider in any judgment of difficulty.  The grammatical structure of a sentence is also important to consider: if the sentence uses a more familiar structure, it is likely to be considered less difficult than a sentence with more unusual structure.  We thus identified two groups of potential features: lexical and grammatical, described below.

\subsection{Lexical Features}

\begin{table*}[]
\centering
\small

\begin{tabular}{|c|c|c|c|c|c|c|c|c|c|}
\hline
\multicolumn{1}{|l|}{} & \multicolumn{3}{c|}{Sentence Only}         & \multicolumn{3}{c|}{In Passage, With Coref} & \multicolumn{3}{c|}{In Passage, No Coref} \\ \cline{2-10} 
Model                  & Acc.             & S.D.   & p-value        & Acc.          & S.D.         & p-value      & Acc.         & S.D.        & p-value      \\ \hline
Oracle (A)             & 90.13\%          & 2.71\% & ---            & 87.81\%       & 1.84\%       & ---          & 87.81\%      & 1.84\%      & ---          \\ \hline
All Features (B)       & \textbf{84.69\%} & 3.46\% & 0.01 $\downarrow$            & \textbf{81.66\%}       & 3.17\%       &  0.005 $\downarrow$            & \textbf{81.91\%}      & 3.27\%      &    $\leftarrow$ 0.04          \\ \hline
AoA + Parse L. (C)     & \textbf{84.33\%} & 3.13\% & 0.001 $\downarrow$ & \textbf{81.27\%}       & 3.93\%       &   0.001 $\downarrow$           & \textbf{80.84\%}      & 3.61\%      &  $\leftarrow$ 0.001            \\ \hline
AoA (D)                & \textbf{79.62\%} & 2.71\% & 0.001 $\downarrow$               & \textbf{79.72\%}       & 2.86\%       &   0.001 $\downarrow$           & \textbf{78.99\%}      & 2.58\%      &    $\leftarrow$ 0.001     \\ \hline
Strat. Random          & 50.28\%          & 1.68\% & ---            & 50.31\%       & 2.01\%       & ---          & 50.31\%      & 2.01\%      & ---          \\ \hline
\end{tabular}
\caption{Mean and standard deviation of accuracy on 200 randomized samples of 20\% held out data. `With coref' indicates coreference features were used. The arrow indicates which immediately adjacent accuracy result is used for p-value comparison, e.g. Model B sentence-only is compared to model C sentence-only, and model B passage, no coref is compared to model B passage, with coref.}
\label{accuracytable}

\end{table*}

For lexical features, based partly on the work of (Song and Cohn 2011) we included the percentage of non-stop words (using NLTK list), the total number of words and the total number of characters as features. 
We included the percentage of words in the text found in the Revised Dale-Chall word list \cite{dalechalllist} to capture the presence of more difficult words in the sentence.

Because sentences that contain rarer sequences of words are likely to be more difficult, and the likelihood of the sentence based on a large corpus should reflect this, we included the n-gram likelihood of each sentence, over each of 1-5 n-grams, as a feature.  The Microsoft WebLM service \cite{msweblm} was used to calculate the n-gram likelihood.  


In the field of psycholinguistics, Age of Acquisition (AoA) refers to the age at which a word is first learned by a child.  A database of 51,715 words collected by \cite{kuperman2012age} provides a rich resource for use in reading difficulty measures. 
With this dataset, we computed several additional features: the average, maximum, and standard deviation of the aggregated AoA for all words in a sentence that were present in the database. Since the data set also includes the number of syllables in each word, and as \cite{kincaid1975derivation} proposes that words with more syllables are more difficult, we also included the average and maximum syllable count as potential features.  

\subsection{Syntactic Features}

We parsed each sentence in the data set using the BLLIP Parser \cite{charniak2005coarse}, which includes a pre-trained model built on the Wall Street Journal Corpus. This provided both a syntactic tree and part of speech tags for the sentence. As Part of Speech tagging is often used as a high-level linguistic feature, we computed percentages for each PoS tag present, since the percentages might vary between difficult sentences and easier sentences. The percentage for each Part of Speech tag is defined as the number of times a certain tag occurred, divided by the total tags. The diversity of part of speech tags was used since this might vary between difficult and easier sentences.

Using the syntactic tree provided by the parser, we obtained the likelihood of the parse, and the likelihood produced by the re-ranker, as syntactic features.  If a sentence parse has a comparatively high likelihood, it is likely to be a more common structure and thus more likely to be easier to read. The length and height of the parse were also included as features, since each of these could reflect the difficulty of the parse. Including the entire parse of the sentence would create too much sparsity since syntactic parses vary highly from sentence to sentence. Therefore, as was done in \cite{Heilman:2008:ASM:1631836.1631845}, subtrees of depth one to three were created from the syntactic parse, and were added as features. This creates a smaller feature set, and one that can potentially model specific grammatical structures that are associated with a specific level of difficulty.

\section{Pairwise Difficulty Prediction Results}


The performance of the logistic regression models trained with different feature sets, for each task, is shown in Table~\ref{accuracytable}.  We reported the mean and standard deviation of the accuracy of each model over 200 randomly selected training and testing splits.  
Each test set consisted of 20\% of the data, and contained 60 aggregate pairs, all of which are sentences (24 in total) that were not present in the training data. The test sets for the sentence-in-passage and sentence-only task contain the same sentence pairs, but the individual judgements are different. 

For comparison, an oracle is included that represents the accuracy a model would achieve if it made the optimal prediction for each aggregate pair.  Due to disagreement within the crowd, the oracle cannot reach 100\% accuracy.  For example, for some pair A and B, if 10 workers selected A as the more difficult sentence, and 4 workers selected B, the oracle's prediction for that pair would be that that A is more difficult.  The judgments of the four workers that selected B would be counted as inaccurate, since the feature set is the same for the judgments with A and the judgments with B. Therefore, the oracle represents the highest accuracy a model can achieve, consistent with the provided labels, using the features provided.  

Examining the results in Table~\ref{accuracytable}, we find the best performing configuration, Model B, used all features as candidates.  The exact number of features selected varied depending on the task.
However, the simplest model, the Age of Acquisition model (D) consisting of the average, standard deviation, and maximum AoA features (sentence-only: 6 features, sentence-in-passage: 12 features) performed well, achieving over 78\% accuracy on all tasks, showing that most of the relative difficulty signal at the sentence level can be captured with a few lexical difficulty features.  The Age of Acquisition + Parse Likelihood model (C) consists of all Age of Acquisition features, plus the likelihood of the parse (sentence-only: 10 features, sentence-in-passage: 20 features)\footnote{The p-value for each accuracy measurement compares its significance, using a paired t-test, to the neighboring model in the direction of the arrow.  For example, the sentence-only Model B is compared to sentence-only Model A.}.

To assess the contribution of different features to the model prediction, feature group importances are reported in Table \ref{cvimportances}.  As features for a given group are often highly correlated with each other, such as in Age of Acquisition, the importance is calculated for feature groups.  Based on the method described for Model B, each feature group is removed from consideration in the model, and the resulting error rate from Model B is used to calculate an importance measure. The most important feature is normalized to have a value of 1.0, with the rest being relative to the difference in error rate from the original model, averaged across splits.

\begin{table}[]

\centering
\scriptsize

\begin{tabular}{|c|c|c|c|}
\hline
\multicolumn{2}{|c|}{Sentence Only} & \multicolumn{2}{c|}{Sentence in Passage (with Coref)} \\ \hline
Feature             & Imp.    & Feature                      & Imp.             \\ \hline
Age of Acq.         & 1.00          & Age of Acq.                  & 1.00                   \\ \hline
Part of Speech      & 0.28          & Syllables                    & 0.27                   \\ \hline
Syn. Score          & 0.22          & Part of Speech               & 0.23                   \\ \hline
Syn. Other          & 0.21          & Syn. Tree                    & 0.18                   \\ \hline
Syllables           & 0.19          & Dale Chall                   & 0.17                   \\ \hline
Ngram L.            & 0.19          & Content Word \%              & 0.17                   \\ \hline
Word Len.           & 0.17          & Word Len.                    & 0.16                   \\ \hline
Dale Chall          & 0.16          & Syn. Other                   & 0.16                   \\ \hline
Content Word \%     & 0.15          & Syn. Score                   & 0.12                   \\ \hline
Syn. Tree           & 0.12          & Ngram L.                     & 0.10                   \\ \hline
\end{tabular}
\caption{Relative feature importance for Model B. Feature importance is the increase in absolute error with a specific feature group removed, averaged across cross-validation folds used for Table 1, and normalized relative to the most informative feature. For Sentence in Passage, feature groups include coreference features.}
\label{cvimportances}

\end{table}
\begin{table}[h!]

\centering
\scriptsize

\begin{tabular}{|l|c|}
\hline
                       & \multicolumn{1}{c|}{Value} \\ \hline
Avg. Abs. Diff         & 9.3                                                    \\ \hline
Avg. Abs. Std Dev      & 7.7                                                    \\ \hline
Pearson's correlation  & \textbf{0.94}*                                 \\ \hline
Spearman's correlation & \textbf{0.94}*                                 \\ \hline
\end{tabular}
\caption{Comparison of rankings generated with and without passage. Asterisk * indicates $p < 0.0001$.}
\label{rankingcomparisontable}

\end{table}

These prediction results show that relative reading difficulty can be predicted for sentence pairs with high accuracy, even with fairly simple feature sets.  In particular, the results for AoA model D, which uses a small number of targeted features, are competitive with the best model B that relies on a much larger feature set. 
The addition of coreference features did result in small but significant changes in the accuracy of the sentence-in-passage task, although in one case the accuracy was reduced with coreference features.  


\section{Ranking Results}

\begin{table}[t!]

\centering
\small

\begin{tabular}{|l|l|l|l|l|}
\hline
\% Diff  & Pearson & p-val. & Spearman & p-val. \\ \hline
Reranker & \textbf{-0.33} & 0.0002 & \textbf{-0.29}  & 0.001 \\ \hline
Parser   & \textbf{-0.33} & 0.0002 & \textbf{-0.28}  & 0.002 \\ \hline
\end{tabular}

\caption{Correlation between difference in rank and percentage difference in features. }
\label{differencecorrelation}

\end{table}



\begin{table*}[!t]
\small
\centering

\begin{tabular}{|l|c|c|c|c|c|c|c|c|}
\hline
                       & \multicolumn{4}{c|}{Sentence}                                                      & \multicolumn{4}{c|}{Sentence-In-Passage}                                           \\ \cline{2-9} 
                       & \multicolumn{2}{c|}{All}                & \multicolumn{2}{c|}{Gold Only}           & \multicolumn{2}{c|}{All}                & \multicolumn{2}{c|}{Gold Only}                 \\ \cline{2-9} 
\multicolumn{1}{|c|}{} & Pearson            & Spearman           & Pearson             & Spearman           & Pearson            & Spearman           & Pearson            & Spearman            \\ \hline
AoA Avg                & \textbf{0.6971}  & \textbf{0.7151}  & \textbf{0.7366}   & \textbf{0.7598}  & \textbf{0.7155}  & \textbf{0.7356}  & \textbf{0.6220}  & \textbf{0.6482}   \\ \hline
AoA Std Dev            & \textbf{0.6366}  & \textbf{0.6596}  & \textbf{0.7074}   & \textbf{0.7385}  & \textbf{0.6779}  & \textbf{0.7023}  & \textbf{0.5742}  & \textbf{0.5825}   \\ \hline
AoA Max                & \textbf{0.7084}  & \textbf{0.6814}  & \textbf{0.8036}   & \textbf{0.7877}  & \textbf{0.7408}  & \textbf{0.7155}  & \textbf{0.6215}  & \textbf{0.6127}   \\ \hline
Parser L.              & \textbf{-0.4942} & \textbf{-0.5297} & \textbf{-0.4605*} & \textbf{0.4920}  & \textbf{-0.4172} & \textbf{-0.4465} & \textbf{-0.5099} & \textbf{-0.5157}  \\ \hline
Reranker L.            & \textbf{-0.4923} & \textbf{-0.5280} & \textbf{-0.4574*} & \textbf{-0.4751} & \textbf{-0.4139} & \textbf{-0.4450} & \textbf{-0.4969} & \textbf{-0.4879*} \\ \hline
\end{tabular}
\caption{Sentence-Only and Sentence-In-Passage Ranking Correlation with Individual Features.  Gold indicates only gold-standard questions were used to build ranking. All correlations have $p < 0.0001$ except those with an asterisk *, which have $p < 0.001$.}
\label{correlationstbl}

\end{table*}

Using the pairwise aggregation method described in Sec.~\ref{rankingcreation}, we ranked sentences by relative difficulty for both sentence-only and sentence-in-passage tasks.  By observing how the overall rank ordering of sentences changes across these conditions, we can identify differences in how workers judged the relative difficulty of sentences with and without context. 

\subsection{Rank Differences}

We report differences in ranking in terms of mean and standard deviation of the absolute difference in rank index of each sentence across the two rankings, along with Pearson's coefficient and Spearman's rank order coefficients.  Comparisons between the rankings for each task are shown in Table \ref{rankingcomparisontable}.

In comparing crowd-generated rankings for the sentence-only and sentence-in-passage task, the results show a statistically significant aggregate difference in how the crowd ranks sentence difficulty with and without the surrounding passage.  While the correlation between the two rankings is high, and the average normalized change in rank position is 7.7\%, multiple sentences exhibited a large change in ranking.  For example, the sentence `\emph{As a result, the police had little incentive to make concessions.}' was ranked significantly easier when presented out of context than when presented in context (rank change: -30 positions). For that example, the surrounding passage explained the complex political environment referred to indirectly in that sentence.

\subsection{Feature Correlation with Rank Differences}

To examine why sentences may be ranked as more or less difficult, depending on the context, we examined the correlation between a sentence's change in rank (Sentence-Only Ranking minus the Sentence-in-Passage ranking) and the normalized difference in feature values between the sentence representation and the remaining context representation. We found that percentage change in parser and reranker likelihoods had the most significant correlation (-0.33) with ranking change, as shown in Table \ref{differencecorrelation}. 

To interpret this result, note that the parser and reranker likelihood represent the probability the parser and reranker models assign to the syntactic parse produced by the sentence.  In other words, they are a measure of how likely it is that the sentence structure occurs, based on the model's training data. If the difficulty of the sentence-in-passage is ranked higher than the sentence alone, this correlates with the target sentence having a syntactic structure with higher likelihood than the average of the surrounding sentence structures.  This means that if a sentence that has a frequently-seen syntactic structure is in a passage with sentences that have less common structures, the sentence within the passage is more likely to be judged as more difficult.  The reverse is also true: if a sentence that has a more unusual syntactic structure is in a passage with sentences with more familiar structures, the sentence without the surrounding passage is more likely to be ranked as more difficult.

We also examined the rank correlation of crowd-generated rankings with rankings produced by sorting sentences based on the value of individual features. In addition to the full rankings, we constructed a ranking produced only by the gold standard examples, denoted \emph{Gold Only} and included this in the comparison. The gold standard questions consist of examples constructed by the authors to have a clear relative difficulty result. The rank correlations are shown in Table \ref{correlationstbl} for both tasks.

The reasons for discrepancies in relative difficulty assessment between the sentence-only and sentence-in-passage conditions require further exploration.  While the correlation between the percentage change in probability of the parse and the difference in ranking is significant, it is not large. It does indicate that despite judges being explicitly told to only consider the sentence, the properties of the surrounding passage may indeed influence the perceived relative difficulty of the sentence.

\subsection{Review of Data}

The pairwise prediction results indicate that a large proportion of the crowdsourced pair orderings can be decided using vocabulary features, due to the strong performance of the Age of Acquisition features.  To identify the relative importance of vocabulary and syntax in our data, we reviewed each pair and judged whether the sentence's syntax or vocabulary, or the combination of both, were needed to correctly predict the more difficult sentence.  For many pairs, either syntax or vocabulary could be used to correctly predict the more difficult sentence since each factor indicated the same sentence was more difficult. 
We found that 19\% of pairs had only a vocabulary distinction, and 65\% of pairs could be judged correctly either by vocabulary or syntax.  Therefore, 84\% of pairs could be judged using vocabulary, which explains the high performance of the Age of Acquisition features.

\begin{table}
\centering
\small

\begin{tabular}{|l|c|c|}
\hline
& \multicolumn{2}{c|}{Crowd}        \\ \cline{2-3} 
         & Pearson     & Spearman   \\ \hline
Expert label & \textbf{0.85}       & \textbf{0.84}       \\ \hline
Document-based label & \textbf{0.70}       & \textbf{0.70}       \\ \hline
\end{tabular}
\caption{Correlation between sentence readability labels and crowd-generated ranking, for expert (sentence-level) and document-based labels (from document readability prediction). All correlations have $p < 0.0001$.}
\label{sentenceleveltable}


\end{table} 

The level of a sentence's source document was used as a proxy for the sentence's grade level when building the pairs.  To build a sentence-level gold standard for this dataset, we asked a teacher with a Master of Education with a Reading Specialist focus and 30 years of experience in elementary and high school reading instruction, to identify the grade level of each sentence.  This expert was asked to assign either a single grade level or a range of levels to each of the 120 sentences.  From this, an expert ranking was created, using the midpoint of each expert-assigned range.  The correlation between the expert sentence ranking and the crowd ranking can be seen in Table \ref{sentenceleveltable}, reinforcing the finding that crowdsourced judgments can provide an accurate ranking of difficulty~\cite{de2014using}.


\section{Conclusion}

Using a rich sentence representation based on lexical and syntactic features leveraged from previous work on document-level readability, we introduced and evaluated several models for predicting the relative reading difficulty of single sentences, with and without surrounding context.  We found that while the best prediction performance was obtained by using all feature classes, simpler representations based on lexical features such as Age of Acquisition norms were  effective.  The accuracy achieved by the best prediction model came within 6\% of the oracle accuracy for both tasks.

Many of the features identified had a high correlation with the rankings produced by the crowd.  This indicates that these features can be used to build a model of sentence difficulty. With the rankings built from crowdsourced judgments on sentence difficulty, small but significant differences were found in how sentences are ranked with and without the surrounding passages.  This result suggests that properties of the surrounding passage of a sentence can change the perceived difficulty of a sentence.

In future work, we plan to increase the number of sentences in our data set, so that additional more fine-grained features might be considered. For example, weights for lexical features could be more accurately estimated with more data. Our use of the crowd-based labels was intended to reduce noise in the ranking analysis, but we also intend to use the pairwise predictions produced by the logistic model as the input to the aggregation model, so that rankings can be obtained for previously unseen sentences in operational settings. Another goal is to obtain absolute difficulty labels for sentences by calibrating ordinal ranges based on the relative ranking. 
Finally, we are interested in the contribution of context in understanding the meaning of an unknown word.

\section*{Acknowledgments}
We thank the anonymous reviewers for their suggestions, and Ann Schumacher for serving as grade level annotator. This work was supported in part by Dept.~of Education grant R305A140647 to the University of Michigan. Any opinions, findings, conclusions or recommendations expressed in this material are the authors', and do not necessarily reflect those of the sponsors.
  
\bibliography{emnlp2016}
\bibliographystyle{emnlp2016}

\end{document}


\maketitle

\section*{Appendix} \label{appendix}

\subsection*{Instructions for the sentence-only task}

Our work concerns helping teachers find texts that are at just the right reading difficulty level for their students. In order to help us , the purpose of this task is to determine the relative reading difficulty of sentences, by comparing a pair of sentences.  In each question, there are two sentences, labelled “Sentence A” and “Sentence B”. Please read both sentences, and then select the one, in your opinion, that would be more challenging to read and understand compared to the other. 

For each sentence, consider which has more difficult vocabulary, that would likely be understood only by a more advanced reader.  For example, the verb in I changed the settings is less difficult to understand than I readjusted the settings. Also, consider the complexity of the grammatical structure of each sentence.   For example, The man is eating a sandwich has a less complex grammatical structure than A sandwich is being eaten by the man.

\textbf{Example 1:}

Sentence A: Each community has many different people who do different things.

Sentence B : Between him and Darcy there was a very steady friendship, in spite of great opposition of character.

All of the words in Sentence A would likely be understood by an elementary school student.  In Sentence B, words such as steady and spite are more advanced words and are likely to be understood by more advanced readers.    The structure of Sentence B is more advanced, as such phrases as in spite of great opposition of character are rarer than phrases in Sentence A.  Therefore, Sentence B is more difficult than Sentence A.

\textbf{Example 2:}

Sentence A: But some photos and films appear to be authentic.

Sentence B : Its buildings were designed  before cars became the standard mode of travel.

In Sentence A,  the word authentic is more advanced, but the remainder of the words are likely to be understood by most readers.  In Sentence B, mode would be considered an advanced word.  Therefore, we can judge the difficulty of the vocabulary to be similar.  However, the structure of Sentence B is more complex than in Sentence A.   Therefore, Sentence B is more difficult than Sentence A.

\textbf{Example 3:}

Sentence A: North Korea frees American helicopter pilot Hall held for 13 days, allowing him to return home to Florida for New Year's Day and keeping alive its nuclear deal with United States.

Sentence B : Whether or no, the mender of roads ran, on the sultry morning, as if for his life, down the hill, knee-high in dust, and never stopped till he got to the fountain.

In Sentence A,  the words nuclear and the name North Korea are advanced. In Sentence B, the words mender and sultry are advanced, and more advanced than the words in Sentence A.  While both sentences have difficult grammatical structures, Sentence B has a high number of commas, which in this case denotes a high number of clauses, which makes it more difficult than Sentence A. Therefore, Sentence B is more difficult than Sentence A.  Note that although Sentence A is longer than Sentence B, Sentence B is more difficult.

\textbf{Example 4:}

Sentence A: Numerous experts agree that this picture may be unique.

Sentence B : All Danny had to do was push some buttons and put the doughnuts on the table.

In Sentence A,  the words experts and numerous are advanced. In Sentence B, there are no words that would be considered advanced.  Therefore, Sentence A is more difficult than Sentence B.  Note that Sentence A is more difficult despite being the shorter of the two sentences.

\subsection*{Instructions for the sentence in passage task}
Our work concerns helping teachers find texts that are at just the right reading difficulty level for their students. In order to help us, the purpose of this task is to determine the relative reading difficulty of sentences, by comparing a pair of sentences. In each question, there are two underlined sentences, labelled “Sentence A” and “Sentence B”, within two passages. Please read the entirety of both passages.  Then, select the underlined sentence that, in your opinion, would be more  challenging to read and understand compared to the other underlined sentence.   Please make your comparison based only on the text of the sentences.

For each sentence, consider which has more difficult vocabulary, that would likely be understood only by a more advanced reader. For example, the verb in I changed the settings is less difficult to understand than I readjusted the settings. Also, consider the complexity of the grammatical structure of each sentence in the passages. For example, The man is eating a sandwich has a less complex grammatical structure than A sandwich is being  eaten by the man.

\textbf{Example 1:}

Sentence A: It all started when Miss Fritz, our fourth grade science teacher, was showing a video about the solar system and different planets. Halfway through the video, I noticed a sparkling metal disc, about the size of a quarter, lying on the floor. \ul{I kept trying to pay attention to the video, but found myself reaching over to grasp the shiny disk that was next to my desk.}  As soon as I touched the metal disk, something strange happened. I wasn’t in the classroom anymore.  I was hovering in the air,  way above the school. I could see the whole town, or rather the rooftops of the whole town. I was a little nervous, but also pretty excited. What was happening? How could I be floating?

Sentence B : Most kids’ backpacks can easily hold school necessities. Is one pack better than another or are they pretty much equal behind the brand name and the price tag? To find out, we bought a half-dozen moderately priced packs plus a messenger bag. \ul{They were all reported to be popular.} We then asked 18 middle-school boys and girls to check them out. We ran lab tests for durability, water-resistance, and other practical stuff to generate the ratings below. The kids didn’t favor one backpack over another. But they quickly made it clear that they preferred a traditional backpack to the messenger bag’s single-strap design.

In Sentence A,  words such as attention and grasp are more difficult words.  The structure of Sentence A is also complex, containing three clauses.  In Sentence B, the words are less difficult, and the sentence structure is relatively simple. Therefore,  Sentence A is more difficult than Sentence B.  Note that the passage of Sentence B contains many other difficult words, such as durability, and is a more difficult passage to read.  Despite this, Sentence A is more difficult than Sentence B when comparing the contents of only the selected sentences.

\textbf{Example 2:}

Sentence A: In 1815 an English banker named Nathan Rothschild made his fortune by relying on messages sent to him by carrier pigeons. English troops were fighting Napoleon’s forces in France, and the English were believed to be losing. A financial panic gripped London. Government bonds were offered at low prices. Few people noticed that Rothschild was snapping up these bonds when everyone else was desperately trying to sell them.  \ul{He knew something others didn't.} A few days later, London learned the truth; the Duke of Wellington had defeated Napoleon at the battle of Waterloo. The value of the bonds soared, and Rothschild became fabulously wealthy ... all because his pigeons had brought him news of the victory before anyone else knew of it.

Sentence B: Some pet birds, such as parrots, can be great talkers. Among the large parrots, the best talkers are African greys and Amazons. The most popular smaller parrots are the budgies, otherwise known as parakeets. One should remember that just because a certain kind of bird can talk does not mean it will talk. Each bird has a different personality. Some birds never learn to talk. Some may learn only a few words or sounds. \ul{Others seem to learn a large vocabulary easily, soaking up new words like some sort of feathered sponge.}  Although each bird is different, younger birds are more likely to learn to talk than older birds. Also, male birds are usually better talkers than females. However, if you teach a bird to whistle before it learns to talk, it may never learn to talk. This might be because whistling is easier for the bird.

In Sentence A, all of the words are at an average level, and the sentence structure is of average complexity.  In Sentence B, there are more difficult words, such as vocabulary, and the sentence structure is more complex than that of Sentence A.  Therefore, Sentence B is more difficult than Sentence A.  Note that some of the topics in Sentence A, financial matters and the Napoleonic wars, which would be considered more advanced do not influence the fact that Sentence B is more difficult.

\textbf{Example 3:}

Sentence A:  He spent months at his easel, often painting into the night, the only light coming from flickering gas lamps. “I have always held that the grandest, most beautiful or wonderful in nature would, in capable hands, make the grandest, most beautiful or wonderful pictures,” the artist later wrote. “If I fail to prove this,I fail to prove myself worthy of the name painter.” \ul{Thomas Moran proved himself more than worthy.} His “Grand Canyon of the Yellowstone,” a monumental seven-by-12-foot oil painting, is one of the finest landscapes in 19th-century American art.  While Moran worked in his studio, Hayden knocked on Congressional doors. With expedition photos and Moran’s vivid field sketches in hand, Hayden had an arsenal of visual ammunition to push forward the park legislation.

Sentence B: The choices in Connecticut and Massachusetts were easy to make, but how did other states choose their trees? The people in Maine and Minnesota picked conifer trees because the trees are used for lumber and shipbuilding, important industries in the two states. Maine chose the white pine, and Minnesota chose the red pine. The people of Alaska named the Sitka spruce as their official state tree. It provided a lightweight wood that served many purposes. \ul{Hawaii is the only state that chose a tree that was not originally from its own state.} The people of Hawaii chose the candlenut tree, which originally came from southeastern Asia. The paste of candlenut kernels was once used to make candles.That is why it is called the candlenut tree. Virginia chose the flowering dogwood as its tree. The flowering dogwood blooms in the spring and grows throughout the state.

In Sentence A, worthy is an advanced word, and the construction of the sentence is more advanced.  In Sentence B, all words would likely be understood by a lower-level reader, and despite the fact that it is longer than Sentence A,  the structure is more common than Sentence B.  Therefore, Sentence A is more difficult than Sentence B.

\textbf{Example 4:}

Sentence A:  On the other hand, seeing a movie in a theater is an experience all its own. For one thing, you can see the movie on a wide screen as the filmmaker intended. \ul{A good film must be changed in someway to make it smaller in order to be viewed on a television screen.}  One way is known as the “pan-and-scan” method, which involves removing some of the  details in the picture, and this results in an image that is not complete. The other way, called “letterboxing,” keeps the image the way it is on the big screen, with one annoying exception: because the big-screen version is wide, the same picture on a television screen must be long and narrow, with black strips above and below it.

Sentence B:  Like all good farmers, Okonkwo had begun to sow with the first rains. He had sown four hundred seeds when the rains and the heat returned. He watched the sky all day for signs of rain clouds and lay awake all night. \ul{Ready at dawn, he returned to his farm to the sight of withering tendrils.} He had tried to protect them from the smoldering earth by making rings of sisal leaves around them. But by the end of the day the sisal rings were burned dry and gray. He changed them everyday, and prayed that the rain might fall in the night. But the drought continued for eight market weeks and the yams were killed.

In Sentence A, there are no difficult words.  In Sentence B, there are several difficult words, such as withering and tendrils.  Sentence A has a relatively simple structure, while Sentence B is more complex, as it has an embedded infinitive. Therefore, Sentence B is more difficult than Sentence A.  Note that this is the case despite Sentence A being longer than Sentence B.